\documentclass[11pt]{article}

\usepackage{amsmath, amsfonts, amssymb}
\usepackage{graphicx,psfrag,epsf}
\usepackage{enumerate}
\usepackage{natbib}
\usepackage{url}
\usepackage[T1]{fontenc}
\usepackage{beramono}
\usepackage{longtable}
\usepackage{float}

\newcommand{\R}{\mathbb{R}}
\newcommand{\N}{\mathbb{N}}

\newcommand{\J}{\mathcal{J}}

\begin{document}
	
\def\spacingset#1{\renewcommand{\baselinestretch}%
{#1}\small\normalsize} \spacingset{1}

  \title{\bf Using Differentiable Programming for Flexible Statistical Modeling}
  \author{Maren Hackenberg, Marlon Grodd, Clemens Kreutz \\
  	Institute of Medical Biometry and Statistics, \\
  	Faculty of Medicine and Medical Center, University of Freiburg, Germany \\
	and \\
	Martina Fischer, Janina Esins, Linus Grabenhenrich \\
	Robert-Koch-Institut, Berlin, Germany \\
	and \\
	Christian Karagiannidis \\
	Department of Pneumology and Critical Care Medicine, Cologne-Merheim Hospital, \\
	ARDS and ECMO Center, Kliniken der Stadt Köln, \\ 
	Witten/Herdecke University Hospital, Cologne, Germany \\
	and \\
    Harald Binder \\
  	Institute of Medical Biometry and Statistics, \\
	Faculty of Medicine and Medical Center, University of Freiburg, Germany \\}
  \maketitle
  \thanks{MH acknowledges funding by the DFG (German Research Foundation) -- 322977937/GRK2344. 
   The authors thank Gerta Rücker for her thoughtful comments on the manuscript.}

\bigskip
\begin{abstract}
	Differentiable programming has recently received much interest as a paradigm that facilitates taking gradients of computer programs. While the corresponding flexible gradient-based optimization approaches so far have been used predominantly for deep learning or enriching the latter with modeling components, we want to demonstrate that they can also be useful for statistical modeling per se, e.g., for quick prototyping when classical maximum likelihood approaches are challenging or not feasible.
	
	In an application from a COVID-19 setting, we utilize differentiable programming to quickly build and optimize a flexible prediction model adapted to the data quality challenges at hand.  
	Specifically, we develop a regression model, inspired by delay differential equations, that can bridge temporal gaps of observations in the central German registry of COVID-19 intensive care cases for predicting future demand. With this exemplary modeling challenge, we illustrate how differentiable programming can enable simple gradient-based optimization of the model by automatic differentiation. 
	This allowed us to quickly prototype a model under time pressure that outperforms simpler benchmark models.	
	We thus exemplify the potential of differentiable programming also outside deep learning applications, to provide more options for flexible applied statistical modeling.
\end{abstract}

\noindent%
{\it Keywords:} Differential Equations; Machine Learning; Optimization; Workflow 
\vfill

\newpage
\spacingset{1.0} 

\section{Introduction}
\label{sec:intro}

Recently, differentiable programming \citep[e.g.,][]{Innes2019_Zygote_dP} has become a prominent concept in the machine learning community, in particular for deep neural networks. Parameter estimation for the latter has long been conceptualized via backpropagation of gradients through a chain of matrix operations. The paradigm of differentiable programming now allows to view artificial neural networks more generally as non-linear functions specified via computer programs, while still providing gradients for parameter estimation via automatic differentiation. For example, this has allowed to 
incorporate other modeling approaches, such as differential equations, that can better reflect the problem structure in the task at hand \citep{Rackauckas2020a}. Given that differentiable programming now enables such a shift of neural networks towards modeling, we conjecture that it might also be useful for enhancing statistical modeling per se. In the following, we describe a modeling challenge, where differentiable programming allowed us to quickly prototype a prediction model for COVID-19 intensive care unit (ICU) data from the central German registry, and also more generally introduce and illustrate differentiable programming for statistical modeling.

While statistical modeling is sometimes considered as distinct from black box algorithmic modeling in machine learning \citep{Breiman2001}, there are many examples where both fields could benefit from the exchange and joint development of techniques \citep[e.g., ][]{Friedman2000}. In particular, \citet{Efron2020} provides a compelling perspective on how pure prediction algorithms relate to traditional methods with respect to the central statistical tasks of prediction, estimation and attribution, and points out directions for combining their respective advantages and thus enriching statistical methods development. In this spirit, we present an illustrative example that integrates concepts of differential equations and differentiable programming with regression modeling. 

Traditional regression-based methods assume a "surface plus noise" formulation \citep{Efron2020}: an underlying structural component describes the true process we are interested in but can only be accessed by noisy observations. Parameter estimation in such models typically is based on maximum likelihood techniques, e.g., using Fisher scoring as an iterative approach. An overview on many different types of corresponding regression modeling approaches is provided, for example, in \citet{FahTut2001}. Machine learning approaches, such as deep neural networks, typically rely on some loss function, which can be based on a likelihood but does not have to. Optimization then is typically performed using a gradient-descent algorithm. In contrast to Fisher scoring, this does not require specification of second derivatives, which makes it more difficult to determine step sizes and quantify uncertainty, but is more generally applicable. In particular, this enables the paradigm of differentiable programming, where loss functions can be specified as computer programs and optimized via automatic differentiation.
While it is a well-established tool in scientific computing \citep[see, e.g.,][]{Griewank1989}, the field of deep learning has only recently more broadly embraced the applicability of automatic differentiation tools to any prediction model that relies on the gradient-based optimization of a differentiable loss function. Furthermore, such frameworks enable to differentiate through computer programs, supporting control flow and recursion \citep{Baydin2017, Innes2019_Zygote_dP}, e.g., allowing to incorporate solvers of differential equations into neural networks to reflect structural assumptions \citep{Chen2018}. Differentiable programming emerged as a descriptive term for this resulting new paradigm \citep{LeCun2018}. Its core idea is to use automatic differentiation to allow for flexible models that seamlessly integrate elements of deep learning and modeling \citep{Rackauckas2020a}.

We are convinced that differentiable programming  can also be useful for applied statistical modeling, e.g., for quick prototyping when other parameter optimization techniques like maximum likelihood Fisher scoring are challenging. Such an approach also more generally has the potential for combining the two optimization paradigms of machine learning and regression-based modeling and thereby opens up new routes for model building. 
To exemplify this potential, we employ differentiable programming to predict prevalent COVID-19 cases in ICUs with regression models, flexibly adapted to handle temporal gaps in longitudinal observations. In particular, we design a statistical model that is inspired by differential equations, introducing dependence of the model on its own past (similar to delay differential equations), and show how differentiable programming allows to solve the corresponding non-linear optimization problem. Note that we use this particular application merely as an example to illustrate why differentiable programming in general provides a useful tool for statistical modeling, without putting too much emphasis on the specific model. 

In the following, we give a brief overview of exemplary applications of differentiable programming in statistical fields and beyond, before outlining the scenario of our specific application. Next, we develop our modeling approach and give an introduction to differentiable programming as a paradigm for optimizing our model and its relation to automatic differentiation and neural networks. We present results on prediction performance, the model sensitivity with respect to time intervals and different model variants, exemplifying the flexibility of such a model fitting process. The closing discussion provides further remarks on the more general applicability of differential programming in statistical modeling. 

\section{Exemplary applications}

\subsection{Differentiable programming in statistics and beyond}

To illustrate how differentiable programming can be applied in more general settings, we briefly present examples both with a statistical focus and from a broader range of fields. 

A key application example of differentiable programming is the integration of differential equations and neural networks to incorporate structural assumptions \citep{Chen2018, Rackauckas2020a}.
For example, such neural differential equations have been used as a flexible alternative for estimating multi-state survival models \citep{Groha2021}. Further, the authors embed neural differential equations into a variational inference framework to quantify the uncertainty of individual cause-specific hazard rates, with gradients for parameter estimation provided by differential programming. More generally, neural ODEs have been integrated into a Bayesian learning framework \citep{Dandekar2020} for uncertainty quantification. This has further been incorporated into non-linear mixed effects model for pharmacometric modeling \citep{Rackauckas2020c}, where differentiable programming allows for jointly optimizing the different model components in an end-to-end framework.

More broadly, differentiable programming has been applied in, e.g., agriculture and environmental science for modeling spread of banana plant disease incorporating climate data \citep{Wang2020} or forecasting reservoir inflows for water ecology management \citep{Zhou2021}. In finance, it enables integration of neural networks and stochastic differential equations, e.g., for an European options book model \citep{Cohen2021}. Further, data-efficient machine learning methods for scientific applications such as physics-informed neural networks \citep{Rackauckas2020b} have been used to bridge the gap between scientific computing and machine learning. More generally, modeling and simulation can be composed by, e.g., differentiable simulation engines for physics \citep{BelbutePeres2018}, robotics \citep{Degrave2018} and quantum computing \citep{Luo2020}. Finally, differentiable programming has been proposed for integrating machine learning-based surrogate models with modeling and simulation in engineering applications \citep{Rackauckas2021} as well as for image processing and rendering  \citep{Li2018a, Li2018b}. 
While covering a diverse range of fields and application scenarios, all these examples share the core element of being built on a differentiable programming framework to obtain gradients for model optimization and parameter estimation, illustrating the broad applicability of this paradigm.

\subsection{A modeling challenge: predicting COVID-19 ICU demand}\label{sec:data}

In the following, we focus on a statistical modeling challenge in a COVID-19 context as an illustrative example of how differentiable programming  can be used to estimate parameters of a statistical model for prediction in a setting where quality issues in the data make straightforward regression modeling challenging. 

At the onset of the pandemic, we had been asked by the Federal Ministry of Health on a short notice to provide a prediction tool for future ICU demand in German hospitals to serve as basis for critical decisions such as transfer of patients between states, or admissions of patients from neighboring countries to German hospitals in the border regions, together with the Robert Koch Institute (RKI).\footnote[1]{\url{https://www.rki.de/DE/Content/InfAZ/N/Neuartiges_Coronavirus/Projekte_RKI/SPoCK.html}} Hence, we had to quickly prototype a model of ICU capacities with prognostic power, based on past daily reports of prevalent cases in hospitals and numbers of new infections in the population. 

During this first wave of COVID-19 infections in Germany, however, many hospitals did not yet report prevalent cases daily, such that the data is characterized by a large amount of missing values, often also over longer periods of time. 
Any prediction modeling approach will have to deal with these missing values. While these could potentially be imputed, any imputation scheme should ideally incorporate typical temporal patterns, which in turn only are obtained after modeling. 
As a further challenge for modeling, in particular during the first wave of infections, the pandemic dynamics differed widely between regions as well as between larger and smaller hospitals, requiring individual models for each hospital. Yet, as data collections had only just been set up, not many time points were available, implying restrictions on the number of parameter that can be estimated in hospital-specific models. 

\begin{figure} 
	\begin{center}
		\includegraphics[width=5in]{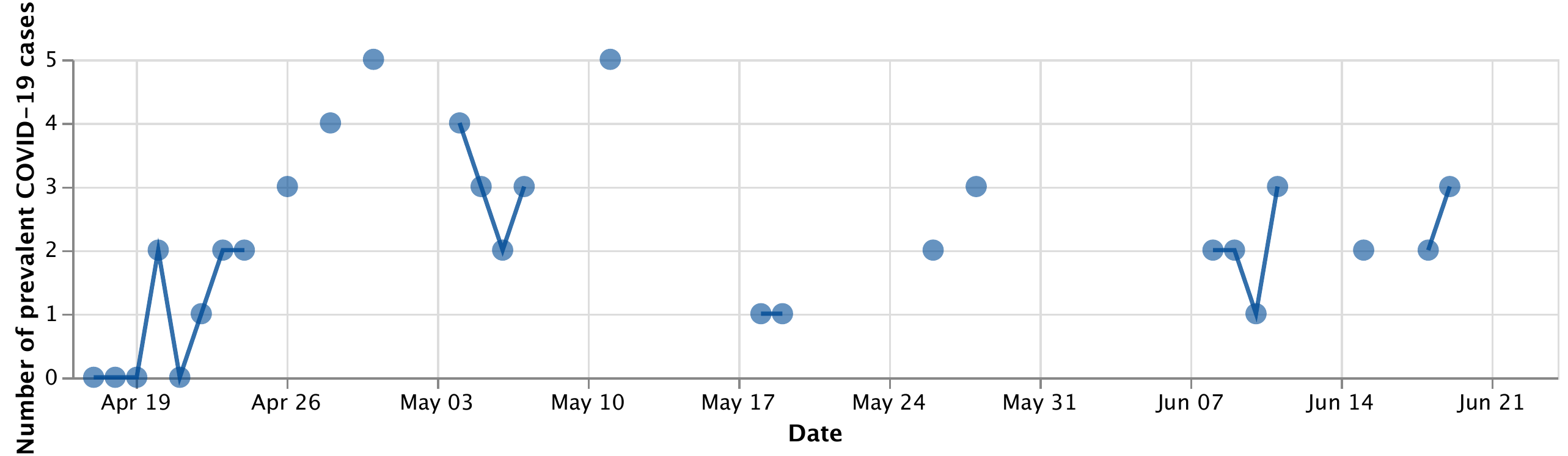}
	\end{center}
	\caption{Exemplary daily numbers of prevalent ICU cases of a hospital from the DIVI registry between April 16, 2020, and June 24, 2020. Reported values are shown in blue.}
	\label{fig:divi-example_cases+imputed}
\end{figure}

Specifically, our model is based on data from the DIVI intensive care registry, a joint project of the German Interdisciplinary Association for Intensive and Emergency Medicine (DIVI) and the RKI that documents capacities for intensive care and records numbers of COVID-19 cases currently treated in ICUs of participating hospitals. 
We consider data from April 16, 2020, onwards, when a regulation came into effect that made daily reporting compulsory for all German acute care hospitals with ICUs. However, many hospitals joined the registry later or did not report on a regular basis, leading to a substantial amount of time intervals without reports: 
Until June 24, 2020, 1281 hospitals have been participating in the registry with a total of $6.41\%$ missing daily reports for prevalent COVID-19 cases. For $463$ ($36.14\%$) hospitals, all daily reports are available, while for $187$ ($ 14.60\%$) hospitals, more than $10\%$ of the daily reports are missing. Additionally, some hospitals have large gaps in their reporting, with e.g. $7.11\%$ of hospitals having more than $25\%$ missing reports, and $10\%$ of hospitals having gaps of five or more consecutive missing reports. 
As an example, Figure~\ref{fig:divi-example_cases+imputed} depicts the daily reports of prevalent COVID-19 ICU cases in a DIVI hospital. 

We observed that larger gaps of consecutive missing reports tend to be associated with a sudden increase or decrease in the numbers of cases after the gap, such that in particular such dynamic phases of increase or decrease tend to be poorly reflected in the data. This shows again that using an imputation approach to capture these missing dynamics is problematic, as, e.g., any form of interpolation will require knowledge of the next report after the gap. However, this is not available when making predictions, which is our primary focus. 

Nonetheless, an approach to explicitly model the course of the prevalent ICU cases over time and to make day-to-day predictions for future demand was urgently needed in the situation outlined above, also as a basis for clinical and political decision making. 
Thus, we developed a model that can flexibly deal with the missingness patterns in the data without requiring imputation, and that can predict future ICU demand by modeling the increments of prevalent ICU cases over time individually for each hospital. 
As we will elaborate further in the next section, here the paradigm of differentiable programming enabled us to quickly prototype a flexible prediction model and estimate its parameters, even though classical maximum likelihood techniques were problematic due to the specific data scenario. 

As further input for predicting the future number of prevalent ICU cases individually for each hospital, we incorporate predicted COVID-19 incidence based on data from the RKI. 
For these predictions, an ordinary differential equation (ODE) model is employed to describe the temporal changes of the number of susceptible, exposed, infected and removed (SEIR) people \citep{Allen2008}. To account for containment measures, a time dependent infection rate is assumed that is described by a smoothing spline \citep{Schelker2012}. All model parameters are estimated by maximum likelihood at the level of the 16 German federal states. The local dynamics at the level of the 412 counties are then estimated by fitting a scaling parameter and an observation error parameter to the reported new infections in each county (see \citet{Refisch2021} for details). The resulting model predictions provide daily information on the local pressure of new cases that could potentially turn into ICU cases. 
We investigate sensitivity of our ICU prediction model to these fitted incidences by comparing results based on the maximum likelihood estimate and the upper and lower bounds of its $95\%$ confidence interval in Section \ref{sec:res_sensitivityanalysis}.

\section{Methods}

\subsection{Developing a flexible regression model to predict the increments of prevalent COVID-19 ICU cases}
\label{sec:methods_model}

Numbers of COVID-19 patients vary between hospitals due to regional differences in infection rates and regulating mechanisms of hospitals within one area (e.g., COVID-19 patients tend to be transferred to larger centers), meaning that predicted county-level COVID-19 incidences translates into a different number of ICU cases for each hospital. Therefore, we fit an individual model for each of the $1281$ participating hospitals from the DIVI register. 
We denote with $y = (y_{1} , \dots, y_{T})^{\top} \in \R^{T}$ the numbers of prevalent ICU cases, where $T \in \N$ is the number of observations days in the registry from April 16, 2020 onwards, and aim at modeling the increments $(y_{t} - y_{t-1})$ for $t = 2,\dots, T$. 

For each hospital, we employ a simple linear regression model with the absolute numbers of prevalent COVID-19 ICU cases $y$ and the incident county-level COVID-19 cases $z = (z_{1} , \dots, z_{T})^{\top} \in \R^{T}$ in the county of the hospital as covariates, representing the potential for decrease (e.g., due to discharge) and increase (through conversion to ICU cases) with respect to the future number of prevalent ICU cases in a hospital, respectively: 
\begin{equation}\label{eq:linearmodel_simple}
(y_{t} - y_{t-1}) = \beta_{1}(t) + \beta_{2}(t) \cdot y_{t-1} + \beta_{3}(t)\cdot z_{t-1}, \quad t=2,\dots, T.
\end{equation} 

To avoid the difficulty of estimating time-dependent parameters in a setting where not many time points were available yet, we decided to model the parameters as constant, i.e., $\beta_i(t) = \beta_i$ for $i=1,2,3$, as a pragmatic solution.

If the numbers of prevalent cases $y$ were a smooth function $y(t) \in \mathcal{C}([t_0, T]; \R)$ defined on the observed time interval, the increments of the prevalent ICU cases could be viewed as a crude approximation to its derivative. Including the absolute numbers of the prevalent ICU cases $y_{t-1}$ at the previous day $t-1$ as a covariate in (\ref{eq:linearmodel_simple}) then resembles a dependence of the approximate derivate on the history of the function itself, like in delay differential equations. Note that this is to be understood as a loose analogy for motivating our approach.

As a consequence, we model continuous-valued increments. While this limits interpretability of predictions as exact discrete ICU utilization numbers, it implies practical benefits: 
When predicting several days into the future, discretizing our predictions would potentially lead to larger errors and less informative results. If we predict increments of, e.g., $0.4$ for $3$ consecutive days, carrying forwards a continuous prediction model results in a predicted increment of $1.2$, while discretizing predictions to counts would result in predicting $0$ new cases. Thus, continuous predictions better allow for gaining tendencies on the upcoming development, which is especially helpful for making decisions on a larger scale, e.g., on transferring patients between states, as was required in our application setting. 

In the following, we describe how this model formulation allows to account for missing daily reports when defining a squared-error loss function for subsequently estimating the model parameters by gradient descent. We encode the information of whether a specific hospital reported their number of prevalent ICU cases at a given time point as a binary vector $r \in \lbrace 0,1 \rbrace^T$, where $r_{t} = 1$ if $y_{t}$ is reported, and $r_{t} = 0$ otherwise, for $t = 1,\dots, T$. 
Whenever the prevalent ICU cases at a certain day have actually been reported, this observation contributes to the loss function, by including the squared difference between the model prediction with the current parameter estimates and the true observed value. However, if a daily report is missing, there is no true value to compare a prediction to, so we use the current state of the model to predict a value for the respective day and move on to the next time point, where the procedure is repeated. In this way, we carry the model prediction forward in time (like a differential equation), until a new observation becomes available.

More formally, defining $\mathrm{d}y_t := y_{t} - y_{t-1}$, the predicted increments of prevalent ICU cases for a specific hospital are given by  
\begin{equation}\label{eq:linearmodel_withmissingobs}
\widehat{\mathrm{d}y}_{t+1} = \beta_1 + \beta_2 \cdot \widetilde{y}_t + \beta_3 \cdot z_{t}, \quad \text{where } \widetilde{y}_t = \begin{cases} 
y_t, & \text{if } r_t = 1, \\
\widetilde{y}_{t-1} + \widehat{\mathrm{d}y}_t, & \text{else.}
\end{cases}
\end{equation}
We can thus define a loss function that includes only the predictions for those time points where a report is available: 
\begin{equation}\label{eq:errorfunction}
\mathrm{L}(y, z, r, \beta) = \frac{1}{\sum_{t=2}^{T} r_t} \sum_{t=2}^{T} r_t \cdot (\widehat{\mathrm{d}y}_t - (\widetilde{y}_t - \widetilde{y}_{t-1}))^2
\end{equation}
Finding optimal model parameters then amounts to minimizing the loss function with respect to the regression coefficients. 

If there are missing daily reports at the beginning of the time interval, i.e., $r_1, \dots, r_s = 0$ for some $s < T$, those values are skipped and the first non-missing value $y_{t_{\mathrm{start}}} = \widetilde{y_{t}}_{\mathrm{start}}, t_{\mathrm{start}} = \min_{t\in \lbrace 1,\dots, T\rbrace}\lbrace r_t \mid r_t = 1\rbrace$ is used to calculate the first prediction $\mathrm{d}y_{t_{\mathrm{start}+1}}$. 

\subsection{Differentiable programming}
\label{sec:methods_dP}

Differentiable programming provides a flexible way to automatically differentiate through any computer program and obtain gradients of the program and its outputs with respect to its parameters. Such a program can consist of any kind of model, e.g., a statistical regression model, a system of differential equations, or a neural network, and a loss function that computes some form of error between the model output and data that the model should be fitted to. If the loss is differentiable as a function of the model parameters, these can be optimized via gradient descent. Here, differentiable programming can be used to calculate these gradients of the loss function with respect to any model parameter in an automated way. The key advantage is that if any model loss function can be written down as a computer program, differentiable programming will allow for estimating the model parameters regardless of how complex the code is (as long as it is differentiable).
This allows to build complex models that can be large and complex, or comprise several distinct building blocks, such as differential equations and neural networks. All these building blocks can then be jointly optimized by combining their outputs into one differentiable loss function and applying automatic differentiation to obtain gradients, using the chain rule to propagate the sensitivities with respect to the parameters backwards through the model components.  

'Differentiable programming' is the term used to describe this new paradigm of model building, where a model can be any computer program that can be conceptualized as a composition of differentiable functions. Exact gradients of the program output with respect to any of its parameters are provided by automatic differentiation, ensuring maximal flexibility and expressivity of the model to be optimized. 
In contrast, manually working out analytical derivatives is time-consuming, prone to errors and not always feasible. Similarly, symbolic differentiation quickly results in long, complex expressions, while also requiring a closed-form expression. Numerical differentiation by finite difference approximations is easy to implement but can incur rounding and truncation errors \citep{Baydin2017}. 

Automatic differentiation overcomes theses limitations by augmenting each variable of the program with the derivative of the program at this variable, and extending all arithmetic operators to this augmented space \citep{Innes2018_Zygote_SSA}. 
In practice, this is realized by a differential operator 
$
\mathcal{J}(f) := x \mapsto (f(x), z \mapsto J_f(x)\cdot z)
$
that maps any function $f$ to a tuple of the value of $f$ at some value $x$, and the evaluation of the Jacobian $J_f(x)$ at some $z$ 
\citep{Innes2019_Zygote_dP}. I.e., it transforms a function to a tuple consisting of the evaluation of both the function and its gradient. With this definition, the gradient of a scalar function $g$ at some $x$ can be obtained by selecting the second element of the tuple $\J(g)(x)$, i.e., the function $z \mapsto J_g(x)\cdot z$, and evaluating it at $z=1$, yielding $\nabla g(x)$.
Specifically, we can use this operator to implement the chain rule. Then, it can be hard coded how $\J$ operates on a set of primitive functions (such as polynomials, $\exp$, $\log$, $\sin$, etc.) and subsequently, differentials of all other functions and complex function compositions can be generated by repeatedly applying the chain rule. 

For example, this strategy is used for fitting neural networks with backpropagation \citep{Rumelhart1986}. Essentially, a neural network is a potentially complex composition of nonlinear functions parameterized by a set of weights and biases, which can be optimized via gradient descent by writing down the model and a loss function as a program, and using automatic differentiation to propagate the gradients of the loss function with respect to all weights and biases backwards through the network.

For our application, we use the automatic differentiation system implemented in the \texttt{Zygote.jl} package \citep{Innes2019_Zygote_dP} for the Julia programming language \citep{Bezanson2017}. A key advantage of Julia with respect to other dynamic languages, such as Python or R, is the broad ecosystem of packages written entirely in Julia itself, making them accessible for automatic differentiation. 
Further, \texttt{Zygote.jl} differs from the majority of state-of-the-art automatic differentiation tools by its use of source-to-source transformation \citep{Innes2018_Zygote_SSA, Innes2019_Zygote_dP} instead of tracing-based mechanisms to evaluate gradients \citep[as in, e.g.,][]{JAX2018, Abadi2016, Innes2018_FluxTracker}. 
While its core idea represents a more general strategy \citep[e.g.,][]{Bischof1996, Pearlmutter2008, Wang2018}, the main novelty of \texttt{Zygote.jl} is its adaptation for a full, high level dynamic language such as Julia \citep{Innes2019_Zygote_dP}. 
The specific characteristics of the Julia implementation are described in detail in \citet{Innes2018_Zygote_SSA}.

As an example, for handling control flow in differentiation many automatic differentiation systems use a tracing approach based on operator overloading: Inputs to the program are wrapped in a new objects, on which any function call does not only produce the result, but also records the operation itself and its inputs. This complete recording includes unrolling all control flow operations. For differentiation, this list is then passed through in reverse order, propagating the gradients backwards through the set of recorded operations. This also limits expressiveness, as it requires re-compilation for every new input value. In contrast, in \texttt{Zygote.jl}, a derivative function is generated directly from the original source code that uses \texttt{goto} instructions to keep control flow intact (see \citet{Innes2018_Zygote_SSA} for details and examples). With this source-to-source transformation \texttt{Zygote.jl} can compile, optimize and re-use a single derivative definition for all input values. It supports control flow, recursion and user defined data types and seamlessly integrates with existing Julia packages. 

\subsection{Fitting the model}
\label{sec:optimisation}

In the following, we illustrate how the parameters of the increment model introduced in Section~\ref{sec:methods_model} can be optimized by differentiable programming. 

Note that the model covariate $\widetilde{y}$ from (\ref{eq:linearmodel_withmissingobs}) depends on the model prediction from previous time steps, such that for any $r_t = 0$,  it holds that 
\begin{align}\label{eq:regmodel_higherorderterms}
\begin{split}
\widehat{\mathrm{d}y}_{t+1} &= \beta_1 + \beta_2 \cdot \widetilde{y}_t + \beta_3 \cdot z_{t} \\
&= \beta_1 + \beta_2 \cdot (\widetilde{y}_{t-1} + \widehat{\mathrm{d}y}_t) + \beta_3 \cdot z_{t} \\
&= \beta_1 + \beta_2 \cdot (\widetilde{y}_{t-1} + (\beta_1 + \beta_2 \cdot \widetilde{y}_{t-1} + \beta_3 \cdot z_{t-1})) + \beta_3 \cdot z_{t},
\end{split}
\end{align}
introducing higher order terms $\beta_2\beta_1, \beta_2^2, \beta_2\beta_3$ of the parameters. 
Thus, we cannot apply Fisher scoring to iteratively obtain a maximum likelihood estimate. Yet, we can find optimal values for the model parameters by using gradient descent on the loss functions, with gradients provided by differential programming, by initializing $\beta^{(0)} = 0$ and updating 
\begin{equation}\label{eq:grad_update}
\beta^{(s)} = \beta^{(s-1)} - \eta \cdot \nabla_{\beta} \mathrm{L}(y, z, r, \beta^{(s-1)}), \quad s = 1, \dots, S, 
\end{equation}
where the number of steps $S \in \N$ and the stepsize $\eta \in \R^3$ are hyperparameters. 
Equation \eqref{eq:regmodel_higherorderterms} shows that the loss as a function of the parameters $\beta$ is essentially a higher-order polynomial, which ensures a sufficiently smooth surface for gradient-based optimization. On the other hand, the higher-order terms in \eqref{eq:regmodel_higherorderterms} imply that the loss is not necessarily convex, such that gradient descent is only guaranteed to find a local optimum. However, by initializing the parameters at $0$ we implicitly use domain knowledge on the generally low numbers of cases. Further, using parameter estimates obtained from past models on less time points as priors for the initialization has been shown to further stabilize and improve the optimization.

To illustrate how even rather complex functions can be straightforwardly optimized with differentiable programming as long as they can be written as computer code, we provide the Julia source code of the loss function, including a loop and control flow elements, in the Supplementary Material, Section 1.

The gradient descent algorithm with the parameter update from (\ref{eq:grad_update}) can then be implemented in just a few lines (see Supplementary Section 1). Here, the exact gradient $\nabla_{\beta} \mathrm{L}$ is obtained as the second element of the output tuple of $\mathcal{J}$ from Section \ref{sec:methods_dP}, i.e., the Jacobian-vector product $z \mapsto J_{\mathrm{L}}(\beta)\cdot z$, evaluated at $z=1$. The loss function can thus be automatically differentiated with respect to the parameters via the \texttt{Zygote.jl} source-to-source differentiable programming framework, without requiring any refactoring.

This flexible optimization of a customized loss function also allows to easily integrate additional constraints and to adapt the model. For example, to prevent overfitting with a smaller number of time points, a $L^2$ regularization term can be added by replacing (\ref{eq:grad_update}) with  
\begin{equation}
\beta^{(s)} = \beta^{(s-1)} - \eta \cdot \nabla_{\beta} (\mathrm{L}(y, z, r, \beta^{(s-1)}) + \lambda \cdot \Vert \beta^{(s-1)} \Vert_2^2), \quad s = 1, \dots, S
\end{equation}
and automatically differentiate through the new loss function for optimization. 

Yet, due to the form of the model predictions (\ref{eq:regmodel_higherorderterms}), estimating the curvature of the loss function is not straightforward, and we cannot use the Fisher information matrix to determine an optimal step size, making the choice of $\eta$ a critical one. 

\section{Results}
\label{sec:res}

\subsection{Implementation}
\label{sec:res_implementation}

We implement the model in the Julia programming language (v1.5.3), using the differentiable programming framework provided by the \texttt{Zygote.jl} package (v0.6.12). Despite our customized, problem-specific model, this implementation allows us to quickly explore model variations and extensions. The complete code to run all analyses and reproduce all tables and figures together with an illustrative Jupyter notebook are available at \url{https://github.com/maren-ha/DifferentiableProgrammingForStatisticalModeling}.

In our gradient-descent algorithm, we set the step size for the update of $\beta_3$ corresponding to the predicted incidences an order of magnitude smaller than the others. By scaling down the incidences, we account for the fact that they are predicted at the level of states which are considerably larger than individual hospital catchment areas, such that only a fraction of new infections in the state will be responsible for potential new cases at one specific hospital in that state. 
On the other hand, the \texttt{Zygote.jl} differentiable programming framework is tightly integrated with the Julia machine learning package \texttt{Flux.jl} \citep{Innes2018_Flux}, which provides a full stack of optimizers for gradient-based learning that determine an optimal step size purely data-based. While some optimizers, in particular versions of the ADAM optimizer \citep{Kingma2015}, also yield better predictions than baseline models, the simple gradient-descent algorithm with manually chosen step size worked well in the present application. We thus decided to incorporate this domain knowledge on inherently different orders of magnitude in the data into our choice of step size, but to otherwise keep the optimization as simple as possible, in a scenario where we were under time pressure to develop a working prototype yet did not have test data available to guide  hyperparameter optimization. 
Rather, with the presented model we aim to showcase an exemplary application scenario where we can benefit from differentiable programming for quick prototyping of a robust statistical prediction model, when no sufficient data or time is available to extensively fine-tune it.

\subsection{Individual models}
\label{sec:res_localmodels}

To evaluate the model, we separately optimize the parameters for each hospital $k=1, \dots, K=1291$, based on all time points available for that hospital except for the last. The prediction for this withheld last time point is then compared to four benchmark models across all hospital in the DIVI registry. 

As a first benchmark, we consider the zero model 
$$\widehat{\mathrm{d}y}_{k,T}^{\mathrm{zero}} := 0, \quad k=1, \dots, K.$$
This benchmark model is motivated by the knowledge that the numbers of COVID-19 infections towards the end of the considered time interval, when the first wave of infections is over, are consistently low. As a second benchmark, we consider a mean model 
$$\widehat{\mathrm{d}y}_{k,T}^{\mathrm{mean}} := \frac{1}{T-2} \sum_{t=2}^{T-1} (y_{k,t} - y_{k,t-1}), \quad k=1, \dots, K,$$
where missing observations are imputed using a last-observation-carried-forward principle. Third, we evaluate a modified version of the mean model 
$$ \widehat{\mathrm{d}y}_{k,T}^{\mathrm{modmean}} := 
\begin{cases} 
0, & \text{if } (y_{k,T-1} - y_{k,T-2}) = 0, \\
\widehat{\mathrm{d}y}_{k,T}^{\mathrm{mean}}, & \text{else,}
\end{cases}  \quad k=1, \dots, K.$$

To account for the parametric complexity of our proposed increment model, as a final benchmark we consider a standard linear regression model with the same covariates, where we impute missing values with a last-observation-carried forward scheme, such that the model can be fitted with standard maximum-likelihood estimation: 

\begin{equation}\label{eq:standardlinreg_locf}
\widehat{\mathrm{d}y}_{t+1}^{\mathrm{linreg}} = \beta_1 + \beta_2 \cdot \widetilde{y}_t + \beta_3 \cdot z_{t}, \quad \text{where } \widetilde{y}_t = \begin{cases} 
y_t, & \text{if } r_t = 1, \\
\widetilde{y}_{t-1}, & \text{else.}
\end{cases}
\end{equation}

When calculating predictions from our proposed model, we have to take hospitals into account where estimation does not converge. 
For the $85$ ($6.6 \%$) hospitals where this occurs, we use the mean model for prediction. Anticipating that the zero models will perform better than the mean models, as they incorporate additional knowledge, this is a conservative choice for evaluating our approach. 

For each of the four models, we sum the squared prediction errors, i.e., the differences to the true increments at the last time point, over all hospitals where the observation at $T$ is not missing, i.e., for all $k$ with $r_{k,T} = 1$: 
\begin{equation}
\begin{split}
	\mathrm{err}^{\mathrm{model}} = \sum_{k=1}^{K} r_{k,T} \cdot (\widehat{\mathrm{d}y}_{k,T}^{\mathrm{model}} &- (y_{k,T} - y_{k,T-1}))^2, \\ &\mathrm{model} \in \lbrace\text{zero, mean, modmean, linreg, increment}\rbrace.
\end{split}
\end{equation}

The results are summarized in Table~\ref{tab:pred_results_globalvslocal}, showing that the increment model optimized via differentiable programming provides the best prediction performance. As expected, the models incorporating knowledge on the large number of zeros perform better than the mean model. Still, our proposal, which uses the mean model as a fallback in cases of non-convergence, outperforms the zero-based benchmark models. 

\subsection{Global vs. individual parameters}
\label{sec:res_globalvslocal}

Additionally, we investigate whether an improvement over individual prediction models for each hospital in the DIVI registry, as presented in Section~\ref{sec:methods_model}, can be achieved by sharing some of the parameters globally, i.e., using the same parameters values for all hospitals. 

To this end, we implement several versions of the model that consider all possible combinations of local and global estimation of individual parameters. 
Technically, we use an approach where in each step individual parameters are fitted for every hospital, but the mean of those individual parameters is taken as a global parameter before proceeding to the next step, such that in step $s$,
\begin{equation}
\beta^{(s,k)} = \beta^{(s-1,\mathrm{mean})} - \eta \cdot \nabla \mathrm{L}(y_k, z_k, r_k, \beta^{(s-1,\mathrm{mean})}), \quad k = 2, \dots, K, 
\end{equation} 
with $\beta^{(s,k)} \neq \beta^{(s,l)}$ for $k, l \in \lbrace 1, \dots, K\rbrace, k \neq l$. After obtaining parameters $\beta^{(s,k)}$ for all $k=1,\dots, K$, we update the global parameter 
\begin{equation}\label{eq:update_mean_beta}
\beta^{(s,\mathrm{mean})} = \frac{1}{K} \sum_{k=1}^{K} \beta^{(s,k)}
\end{equation}
and proceed to the next step $s+1$. 

To explore different combinations of local and global parameters, the update (\ref{eq:update_mean_beta}) can be applied only in specified dimensions such that for any given subset $I\subset \lbrace 1, 2, 3 \rbrace$, $\beta_I^{(s,\mathrm{mean})} = \frac{1}{K} \sum_{k=1}^{K} \beta_I^{(s,k)}$ while for all $j \notin I$, $\beta_j^{(s,k)}$ remains unchanged. 

\begin{table}[]
	\caption{Prediction performance of benchmark models and the proposed increment model with different combinations of global and individual parameters.
	\label{tab:pred_results_globalvslocal}}
	\begin{center}
	\begin{tabular}{llllll}
		\textbf{Prediction model} & \multicolumn{5}{l}{\textbf{Squared error}}   \\
		& Sum & Mean  & 1st Quartile & Median & 3rd Quartile \\ \hline
		Zero model               & 86.0               & 0.07      & 0.0 & 0.0 & 0.0     \\
		Mean model               & 89.239   & 0.072     & 0.0 & 0.0 & 0.002     \\
		Modified mean model      & 84.342  & 0.069     & 0.0 & 0.0 & 0.0      \\
		Linear regression   & 181.068 & 0.147      & 0.0 & 0.001 & 0.015  \\ \hline
		\multicolumn{5}{l}{Increment model with \textbf{global} $\mathbf{\beta_2}$ for prevalent cases} \\ \hline
		global $\beta_1$; individual $\beta_3$ & 98.833  & 0.08  & 0.0 & 0.0 & 0.001  \\
		individual $\beta_1$; individual $\beta_3$ & 94.566  & 0.077  & 0.0 & 0.0 & 0.001 \\
		global $\beta_1$; global $\beta_3$ & 90.406  & 0.073  & 0.0 & 0.0 & 0.0      \\
		individual $\beta_1$; global $\beta_3$ & 88.154  & 0.072  & 0.0 & 0.0 & 0.0  \\  \hline 
		\multicolumn{5}{l}{Increment model with \textbf{individual} $\mathbf{\beta_2}$ for prevalent cases} \\ \hline
		global $\beta_1$; global $\beta_3$ & 78.976  & 0.064  & 0.0 & 0.0 & 0.0   \\
		individual $\beta_1$; global $\beta_3$ & 77.652  & 0.063  & 0.0 & 0.0 & 0.0     \\
		global $\beta_1$; individual $\beta_3$ & 75.712  & 0.062   & 0.0 & 0.0 & 0.002    \\
		individual $\beta_1$; individual $\beta_3$ & \textbf{74.746}  & 0.061 & 0.0 & 0.0 & 0.001
	\end{tabular}
	\end{center}
\end{table}

From the results summarized in Table~\ref{tab:pred_results_globalvslocal}, it can be seen that the model with all individual parameters still performs best. However, individual estimation of the parameter for the prevalent cases, which reflects internal processes of the hospital, more strongly improves the prediction performance than individual estimation of the intercept parameter or of the parameter for the incidences, representing an external pressure. 

The distribution of the squared errors across hospitals is highly skewed, with a prediction error of close to zero for many hospitals, but some hospitals with very large errors. However, these hospitals are not systematically mis-predicted by a specific model, but have high errors across all prediction models. For example, sudden jumps in the number of cases on the day held out for prediction cannot be captured based on the previous data seen by the models.  

Notably, imputing missing reports with last observation carried forward and using standard maximum likelihood estimation performs worse than the simpler baselines. Further, exemplary analyses for individual hospitals show that the our proposal of flexibly bridging gaps of missing reports provides lower minima of the corresponding loss function (see Supplementary Figure 1). 

\subsection{Sensitivity Analysis}
\label{sec:res_sensitivityanalysis}

To investigate the sensitivity of the model with respect to the predicted incidence data used as a covariate, we repeated the analysis presented in the previous Section \ref{sec:res_globalvslocal} with the lower and upper bound of the confidence interval of the maximum likelihood estimate for the predicted incidences (see \citet{Refisch2021} for details on the underlying model and estimation). We found similar results to those based on the maximum likelihood presented in Table \ref{tab:pred_results_globalvslocal} above, with the upper bound yielding slightly worse and the lower bound yielding slightly better predictions (see Supplementary Material, Section 3.1 for details).

The proposed increment model is based on the assumption that the parameters for a hospital are constant in the course of time. To investigate how critical this assumption is, we perform a sensitivity analysis by using different temporal subsets of the data for model fitting, which is also useful for judging variability and potential bias more generally. Specifically, we construct time intervals that are only half as long as the available observation interval of length $70$, starting at each day $t_{\mathrm{start}} = 1, \dots, 36$ one after another, and defining the intervals $I_{t_{\mathrm{start}}} = [t_{\mathrm{start}}, t_{\mathrm{start}} + 34]$, such that $I_1 = [1, 35], I_2 = [2, 36], \dots, I_{36} = [36, 70]$. On each interval, we evaluate the prediction of the baseline models defined in Section~\ref{sec:res_localmodels} and fit models for all combinations of individual and global parameters, as in Section~\ref{sec:res_globalvslocal}. 

Since at the beginning of the observation period, numbers of prevalent cases are globally higher and then decrease until mainly zeros dominate the dataset, the prediction errors for the earlier time intervals are also generally higher. Thus, the absolute values of prediction errors are not directly comparable between different time intervals.  For a first overview of the model behavior across different time intervals, we therefore evaluate the relative improvement of the model over the mean model, i.e., the difference in prediction error $(\text{err}_{\text{mean}} - \text{err}_{\text{increment}})$. Larger values correspond a smaller error of the proposed model compared to the mean model and thus better prediction performance. In Table~\ref{tab:quantiles_results_globalvslocal_resamplings}, we report the first and third quartile as well as the median of the results over all $36$ time intervals. While we present the relative improvement over the mean model as the best benchmark that does not incorporate knowledge on the large number of zeros, the relative improvement over all other benchmark models is presented in the Supplementary Tables 3, 4 and 5. 

\begin{table}
	\caption{Quantiles of improvement in prediction error of the proposed increment model over the mean model, obtained across all time intervals half as long as the entire time period. 
	Larger values imply a smaller error of the increment model compared to the mean model and thus better prediction performance.
	\label{tab:quantiles_results_globalvslocal_resamplings}}
	\begin{center}
		\begin{tabular}{lccc}
			\textbf{Increment model parameters} & \multicolumn{3}{c}{\textbf{Differences of squared error}} \\
			& \multicolumn{3}{c}{(mean model - increment model)} \\ 
			&  1st quartile & median & 3rd quartile \\  \hline
			\textbf{global parameter $\beta_2$ for prevalent cases} & & & \\ \hline
			global $\beta_1$; global $\beta_3$ & 1.997 & 8.940 & 15.292 \\
			individual $\beta_1$; global $\beta_3$  & 5.720 & 12.349 & 16.649 \\
			global $\beta_1$; individual $\beta_3$ & -14.627 & 4.050 & 10.132 \\
			individual $\beta_1$; individual $\beta_3$;  & 1.346 & 7.431 & 14.669 \\ \hline
			\textbf{individual parameter $\beta_2$ for prevalent cases} & & & \\ \hline
			global $\beta_1$; global $\beta_3$ & 7.399 & 14.182 & 18.106 \\
			individual $\beta_1$; global $\beta_3$ & 7.869 & 15.112 & 22.147 \\
			global $\beta_1$; individual $\beta_3$ & 0.760 & 11.590 & 19.399 \\
			individual $\beta_1$; individual $\beta_3$;  & 1.997 & 10.921 & 19.666 \\ \hline			 
		\end{tabular}
	\end{center}
\end{table}

For most combinations, the proposed increment model yields a better prediction performance than the simple mean model on the majority of data subsets, with only one combination with a large negative 1st quartile. As in Section~\ref{sec:res_globalvslocal}, we also generally observe a better prediction performance of the increment model compared to the mean model for all model versions with an individual estimation of the parameter for the prevalent cases. Overall, the results from the previous sections generalize to models fitted only on subsets of the data, indicating general robustness of our model and optimization method. 

Additionally, we take a look at exemplary predictions of the proposed model for larger gaps of missing daily reports that originally motivated our method development. Again, we evaluate these predictions based on the data from the entire time span and all data subsets from the $36$ shorter time intervals and for different combinations of individual and global parameters. 
We exemplarily illustrate the results for the hospital in the DIVI register depicted in Figure~\ref{fig:divi-example_cases+imputed} from Section~\ref{sec:data} with its larger gaps in daily reporting. 

\begin{figure} 
	\begin{center}
		\includegraphics[width=\textwidth]{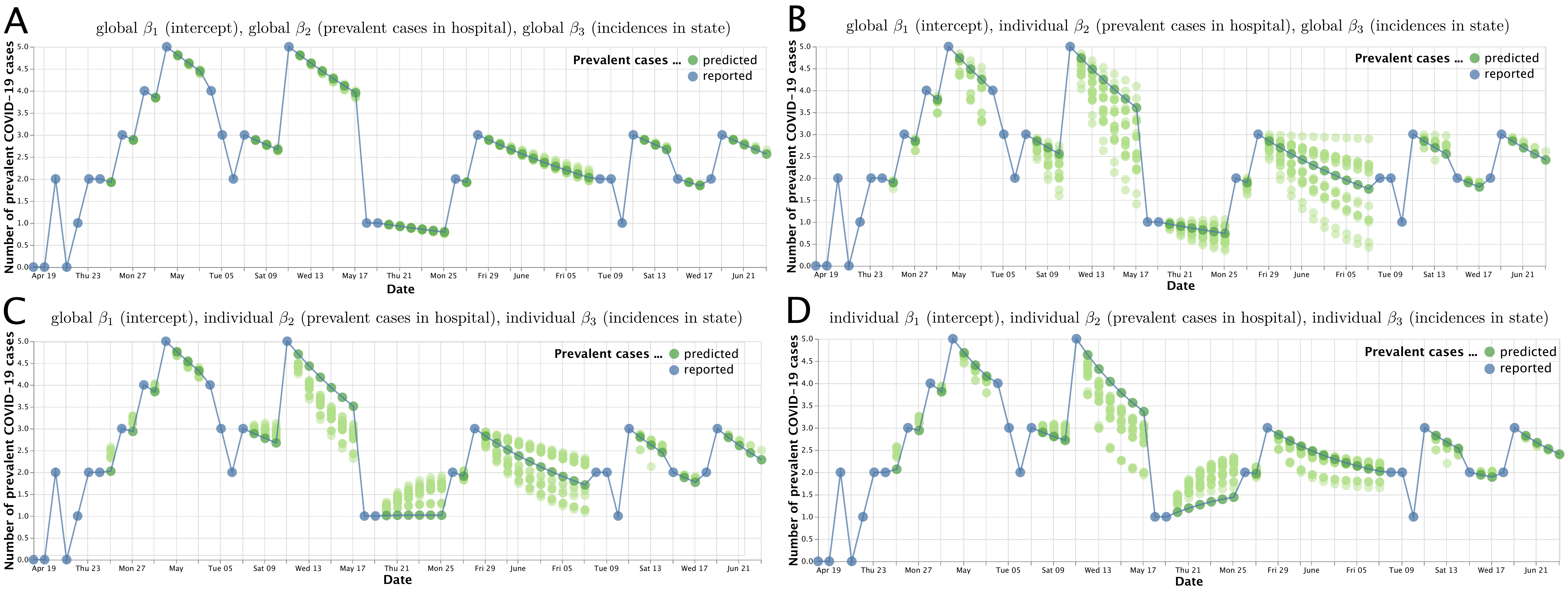}
	\end{center}
	
	\caption{Prevalent cases of one DIVI hospital with observed values shown in blue and model predictions for the missing observations shown in green. The panels depict four combinations of global and individual parameters of the proposed model. In each panel, in addition to the observed values (blue dots connected with blue line), we show both the predicted values based on the respective model optimized on the entire dataset (dark green dots connected with blue line) and the predictions based on all $36$ data subsets obtained from the shorter time intervals (lighter green dots).}
	\label{fig:panel_localglobal}
\end{figure}

In Figure~\ref{fig:panel_localglobal}, we show exemplary predictions for missing reports of prevalent COVID-19 cases obtained from the increment model with different combinations of individual and global parameters. In the models corresponding to panels A-C, the intercept parameter is estimated globally. Panel A represents the all-global model, while in panel B, only the parameter for the absolute value of the prevalent cases is estimated individually, thus corresponding to a model with individually estimated internal processes and globally estimated external pressure. In panel C, only the intercept parameter is estimated globally.

Note that the model is developed for a prediction setting, i.e., linearly interpolating between reported data is not feasible for our purpose, as we do not know a priori what the next value after a gap of missing reports will be when making predictions after one or more days of missing reports. 

As a first observation, the variance of the predictions based on the data subsets (indicated by how wide the light green dots spread out) increases when parameters are estimated individually, as would be expected. On the other hand, global estimation can introduce a bias, as can, e.g., be seen from the predictions for missing values between May, 20 and May, 25: 
Based on the observations before and after that gap, the true development in this period has to be an increase in prevalent cases. Increases are related to higher external pressure due to more incident cases in the population and are thus modeled by the parameter $\beta_3$. During that particular period, numbers of cases are however globally falling, and the increase in prevalent cases is specific to the considered hospital. Thus, global estimation of the corresponding parameter fails to model that development, introducing a bias in panels A and B. When estimating that parameter individually, however, the increase in prevalent cases is captured, as can be seen in panels C and D.

Finally, the trajectory of the exemplary hospital exhibits distinct phases of rising and falling numbers of cases. This more generally applies to the overall development across hospitals, implying that our modeling assumption of constant parameters may be (overly) simplifying and time-dependent parameters would be better suited to capture these phases. 

As there is no ground truth for the missing reports of the hospital in Figure \ref{fig:panel_localglobal}, we further evaluate the capability of our proposed model to bridge gaps of missing reports by randomly censoring reports from hospitals with complete reporting, and assessing how well they are recovered. Specifically, we consider all hospitals with no missing reports ($463$ in total) and evaluate the performance of all models in recovering reports that were censored at random at rater of $10\%$, $25\%$, $50\%$ and $75\%$. In all scenarios the increment model provides the lowest median recovery error across hospitals, and the gap in prediction performance increases for higher proportions of censoring, indicating that our proposed model is particularly suited to deal with larger gaps in reporting (for details, see Supplementary Table 6).

\section{Discussion}
\label{sec:discussion}

Differentiable programming is part of a current paradigm shift in the field of deep neural networks towards incorporating model components, such as differential equations, that better formalize knowledge on the problem at hand. We described an exemplary application of differentiable programming for statistical modeling per se, to illustrate how such techniques could also be useful in applied statistical work, e.g., for quick prototyping. In an exemplary scenario in a COVID-19 setting, characterized by a substantial amount of missing daily reports, we have adapted a regression model for the increments of prevalent ICU cases to flexibly handle missing observations in the data, inspired by differential equations. Using a differentiable programming framework allowed for tailoring the model to the specific requirements of the data, yet optimizing it efficiently.

Specifically, we have shown that our modeling strategy allows for more accurate predictions on the given dataset compared to simple benchmark models. Moreover, the differentiable programming framework provided a convenient tool to quickly explore variations of the model with respect to global vs. individual estimation of parameters and to investigate factors predominantly influencing prediction performance. It thus facilitated exploring the effect of model components and ultimately a more in-depth understanding of the problem.

Using data subsets defined by sliding time intervals for model fitting, we have investigated the robustness of our method and optimization procedure and have found our results with respect to prediction performance and global vs. individual parameter estimation to roughly generalize across subsets. Additionally, we have exemplarily illustrated the ability of the model to bridge gaps of missing observations and have discussed the effect of different global and individual parameters estimates on the prediction bias and variance.

With respect to the application setting of the first COVID-19 wave, the dynamic situation of the pandemic exemplifies a scenario where we can benefit from differentiable programming as it allowed us to quickly prototype a model under time pressure and flexibly adapt it to the challenging data scenario with many missing values. Such an approach can be more generally useful in future epidemic scenarios, when data collection is still ongoing and incomplete, yet interpretable models with predictive capabilities that can be flexibly adapted to the challenge at hand are urgently needed. 

While our models with different combinations of global and individual parameters generally allow for an intuitive interpretation and the identification of external pressure and internal processes for individual hospitals, time-dependent parameters could be more suitable to accurately capture distinct phases of development in the data, which yet has to be addressed in future research. In its current form, the assumption of time-constant parameters can introduce bias that may lead to underestimation of overall less prominent trends. Subsetting the data with a sliding window approach as in Section \ref{sec:res_sensitivityanalysis} and interpolating the resulting parameters would be a first step in that direction. 
Furthermore, the discussed mixtures of individual and global parameters currently rely on a coarse procedure of taking a global mean across all hospitals after each step. For further model refinement, groups of hospitals could be modeled together to uncover cluster structures within the dataset and identify groups of hospitals sharing common developments. Such similarity could be determined either based on the regression parameters or the predicted trajectories, and potentially also take into account structural characteristics of hospitals. 
Despite these present limitations, our method could also be more broadly applicable to general time-series data with missing values from an incomplete data collection. 

Yet, our main aim was to provide an example of how differentiable programming allows to flexibly combine different tools and modeling strategies led by the requirements of a challenging modeling task, and to showcase model optimization with a gradient-based approach in an easy-to-implement and efficient way, rather than the specific modeling example. Differentiable programming can thus aid statistical modeling where established approaches, such as maximum likelihood Fisher scoring, are not applicable and prove beneficial particularly in application-driven settings with a focus on prediction performance.

In short, integrating differentiable programming into statistical modeling can enrich the statistician's toolbox and be a step towards uniting the estimation-focused data modeling and the prediction-oriented algorithmic culture and their respective tools, allowing the fields to interact and mutually inspire each other. 

\bibliographystyle{agsm}
\bibliography{Revision_bibfile}

\pagebreak

\begin{center}
	\textbf{\Large Supplementary Tables and Figures for the Manuscript 
		\\ Using Differentiable Programming for Flexible Statistical Modeling}
\end{center}

\setcounter{equation}{0}
\setcounter{figure}{0}
\setcounter{table}{0}
\setcounter{page}{1}
\setcounter{section}{0}
\makeatletter
\renewcommand{\theequation}{S\arabic{equation}}
\renewcommand{\thefigure}{S\arabic{figure}}
\renewcommand{\thesection}{S\arabic{section}}
\renewcommand{\bibnumfmt}[1]{[S#1]}
\renewcommand{\citenumfont}[1]{S#1}

\section{Original Julia code of the increment model loss function and gradient descent optimization}
\label{sec:supp_lossfun}

Original Julia implementation of the loss function of our proposed model, including a loop and control flow elements. 

\begin{minipage}[t]{\textwidth}
	\setlength\parindent{15pt}\fussy
	\linespread{1.1}\footnotesize
	\texttt{\textbf{function} loss (y, z, r, beta) \\
		\indent \indent sqerror = 0.0 \textit{\# squared error} \\
		\indent \indent firstseen = false \textit{\# set to true after skipping potential missings} \\
		\indent \indent last\_y = 0.0 \textit{\# prevalent cases from previous time point} \\
		\indent \indent contribno = 0.0 \textit{\# number of non-missing observations} \\
		\indent \indent \textbf{for} t = 1:(length(y)) \\
		\indent \indent \indent \textit{\# skip missings at the start until first reported value} \\
		\indent \indent \indent \textbf{if} !firstseen \\
		\indent \indent \indent \indent \textbf{if} r[t] == 1 \\
		\indent \indent \indent \indent \indent firstseen = \textbf{true} \\
		\indent \indent \indent \indent \indent last\_y = y[t] \\
		\indent \indent \indent \indent \textbf{else} \\
		\indent \indent \indent \indent \indent \textbf{continue} \\
		\indent \indent \indent \indent \textbf{end} \\
		\indent \indent \indent \textbf{else} \textit{\# make a prediction for the current increment} \\
		\indent \indent \indent \indent pred\_dy = beta[1] + beta[2] * last\_y + beta[3] * z[t-1] \\
		\indent \indent \indent \indent \textbf{if} r[t] == 1 \\
		\indent \indent \indent \indent \indent dy = y[t] - last\_y \\
		\indent \indent \indent \indent \indent sqerror += (dy - pred\_dy)\^{}2 \\
		\indent \indent \indent \indent \indent contribno += 1.0 \\
		\indent \indent \indent \indent \indent last\_y = y[t] \\
		\indent \indent \indent \indent \textbf{else} \\
		\indent \indent \indent \indent \indent last\_y += pred\_dy \\
		\indent \indent \indent \indent \textbf{end} \\
		\indent \indent \indent \textbf{end} \\
		\indent \indent \textbf{end} \\
		\indent \indent \textbf{return} sqerror/contribno \textit{\# return MSE over all reported time points} \\
		\indent \textbf{end}
	} \\
\end{minipage}

The gradient descent algorithm with the parameter update from Equation (5) can then be implemented in just a few lines:

\begin{minipage}[t]{\textwidth}
	\setlength\parindent{15pt}\fussy
	\linespread{1.1}\footnotesize
	\texttt{beta = [0.0; 0.0; 0.0] \\
		\indent \textbf{for} steps = 1:nsteps \\
		\indent \indent gradloss = gradient(arg -> loss(y, z, r, arg), beta)[1] \\
		\indent \indent beta .-= eta .* gradloss \\
		\indent \textbf{end}
	}\\
\end{minipage}

\section{Comparison of loss curves based on last observation carried forward vs. bridging gaps without imputation}
\label{sec:supp_losscurve_comparison}

\begin{figure}[H]
	\begin{center}
		\includegraphics[width=\textwidth]{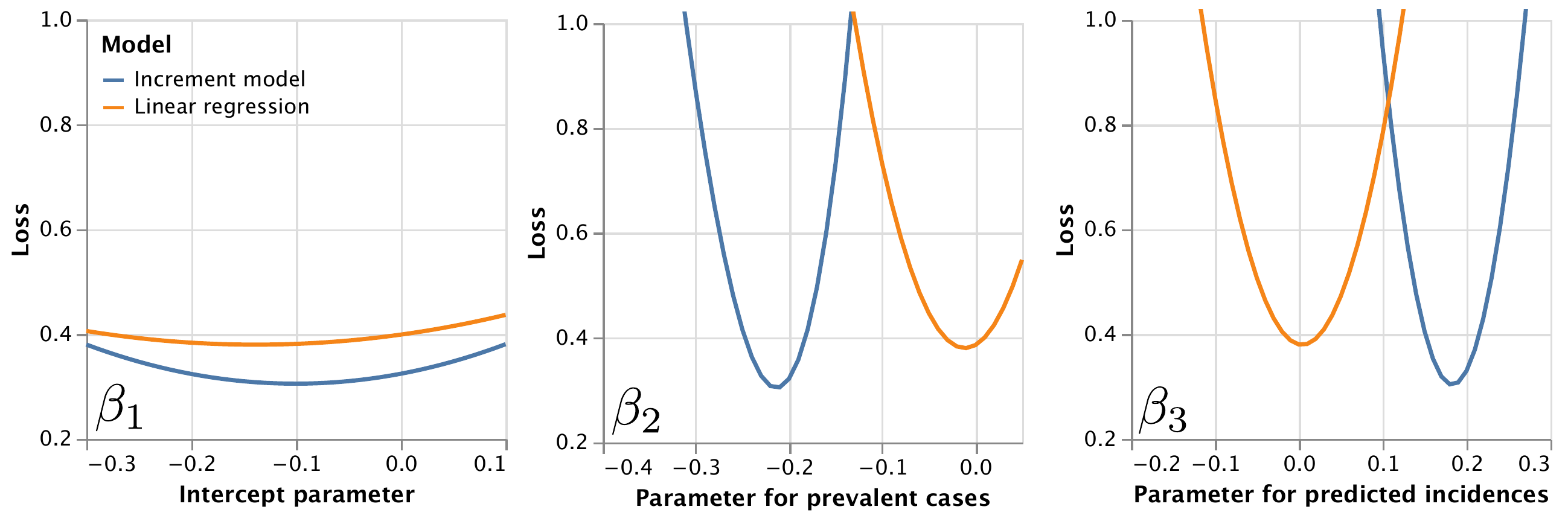}
	\end{center}
	\caption{Loss curves for a least-squares loss from a classical linear regression model (corresponding to standard maximum likelihood estimation) based on last-observation-carried-forward imputation (shown in orange) versus our proposal of flexibly bridging gaps of missing reports without imputation and a differentiable-programming-based optimization (shown in blue) for one exemplary hospital from the DIVI registry. In each panel, the x-axis corresponds to one of the three coefficients to be estimated and the y-axis to the corresponding loss values.}
	\label{fig:supp_losscurves}
\end{figure}

\newpage

\section{Sensitivity Analysis}

\subsection{Using the upper and lower bound of the confidence interval for the maximum likelihood estimate of the predicted incidences used as a covariate }

\begin{table}[h]
	\caption{Prediction performance of benchmark models and the proposed increment model with different combinations of global and individual parameters, where the upper bound of the confidence interval for the maximum likelihood estimate of the predicted incidences was used as a covariate in the model.}
	\label{tab:supp_allcombinations_ub}
	\begin{tabular}{lccccc}
		\textbf{Prediction model} & \textbf{Squared error} &       &              &        &              \\
		& Sum                    & Mean  & 1st Quartile & Median & 3rd Quartile \\ \hline 
		Zero model                & 86.0                   & 0.07  & 0.0          & 0.0    & 0.0          \\
		Mean model                & 89.239                 & 0.072 & 0.0          & 0.0    & 0.002        \\
		Modified mean model       & 84.342                 & 0.069 & 0.0          & 0.0    & 0.0          \\
		Linear regression         & 173.980                & 0.141 & 0.0          & 0.001  & 0.016        \\ \hline
		\multicolumn{6}{l}{Increment model with \textbf{global} $\mathbf{\beta_2}$ for prevalent cases}                           \\ \hline
		global $\beta_1$; individual $\beta_3$  & 97.836                 & 0.079 & 0.0          & 0.0    & 0.001        \\
		individual $\beta_1$; individual $\beta_3$  & 96.349                 & 0.078 & 0.0          & 0.0    & 0.001        \\
		global $\beta_1$; global $\beta_3$  & 90.629                 & 0.074 & 0.0          & 0.0    & 0.0          \\
		individual $\beta_1$; global $\beta_3$  & 88.308                 & 0.072 & 0.0          & 0.0    & 0.0          \\ \hline 
		\multicolumn{6}{l}{Increment model with \textbf{individual} $\mathbf{\beta_2}$ for prevalent cases}            \\ \hline
		individual $\beta_1$; global $\beta_3$  & 79.310                 & 0.064 & 0.0          & 0.0    & 0.0          \\
		individual $\beta_1$; individual $\beta_3$ & 76.940                 & 0.063 & 0.0          & 0.0    & 0.002        \\
		global $\beta_1$; individual $\beta_3$  & 76.046                 & 0.062 & 0.0          & 0.0    & 0.003        \\
		global $\beta_1$; global $\beta_3$  & 75.839                 & 0.062 & 0.0          & 0.0    & 0.0         
	\end{tabular}
\end{table}

\begin{table}[h]
	\caption{Prediction performance of benchmark models and the proposed increment model with different combinations of global and individual parameters, where the lower bound of the confidence interval for the maximum likelihood estimate of the predicted incidences was used as a covariate in the model.}
	\label{tab:supp_allcombinations_lb}
	\begin{tabular}{lccccc}
		\textbf{Prediction model} & \textbf{Squared error} &       &              &        &              \\
		& Sum                    & Mean  & 1st Quartile & Median & 3rd Quartile \\ \hline 
		Mean model                & 89.239                 & 0.072 & 0.0          & 0.0    & 0.002        \\
		Modified mean model       & 84.342                 & 0.069 & 0.0          & 0.0    & 0.0          \\
		Zero model                & 86.0                   & 0.07  & 0.0          & 0.0    & 0.0          \\
		Linear regression         & 177.554                & 0.144 & 0.0          & 0.001  & 0.013        \\ \hline
		\multicolumn{6}{l}{Increment model with \textbf{global} $\mathbf{\beta_2}$ for prevalent cases}            \\ \hline 
		global $\beta_1$; individual $\beta_3$   & 93.414                 & 0.076 & 0.0          & 0.0    & 0.0          \\
		global $\beta_1$; global $\beta_3$   & 89.979                 & 0.073 & 0.0          & 0.0    & 0.0          \\
		individual $\beta_1$; individual $\beta_3$   & 89.866                 & 0.073 & 0.0          & 0.0    & 0.0          \\
		individual $\beta_1$; global $\beta_3$   & 87.715                 & 0.071 & 0.0          & 0.0    & 0.0          \\ \hline
		\multicolumn{6}{l}{Increment model with \textbf{individual} $\mathbf{\beta_2}$ for prevalent cases}            \\ \hline 
		individual $\beta_1$; global $\beta_3$   & 77.447                 & 0.063 & 0.0          & 0.0    & 0.0          \\
		global $\beta_1$; global $\beta_3$   & 74.714                 & 0.061 & 0.0          & 0.0    & 0.0          \\
		global $\beta_1$; individual $\beta_3$   & 72.835                 & 0.059 & 0.0          & 0.0    & 0.0          \\
		individual $\beta_1$; individual $\beta_3$   & 72.751                 & 0.059 & 0.0          & 0.0    & 0.0         
	\end{tabular}
\end{table}

\newpage 

\subsection{Relative improvement of the increment model over benchmark models across subsets based on shorter time intervals}

\subsubsection{Relative improvement over the zero model}

\begin{table}[H]
	\caption{Quantiles of improvement in prediction error of the proposed increment model over the modified mean model, obtained across all time intervals half as long as the entire time period.}
	\label{tab:sup_quantiles_resamples_modmean}
	\begin{tabular}{lccc}
		\textbf{Increment model parameters}         & \multicolumn{3}{c}{\textbf{Differences of squared error}}            \\
		& \multicolumn{3}{c}{(modified mean model - increment model)} \\
		& 1st Quartile         & Median         & 3rd Quartile        \\ \hline 
		\multicolumn{4}{l}{\textbf{global parameter} $\mathbf{\beta_2}$ \textbf{for prevalent cases}}     \\ \hline 
		global $\beta_1$; individual $\beta_3$         & -18.677              & 0.364          & 3.982               \\
		individual $\beta_1$; individual $\beta_3$    & -2.609               & 3.825          & 8.489               \\
		global $\beta_1$; global             & -0.307               & 4.116          & 8.985               \\
		individual $\beta_1$; global         & 2.619                & 6.289          & 10.211              \\ \hline 
		\multicolumn{4}{l}{\textbf{individual parameter} $\mathbf{\beta_2}$ \textbf{for prevalent cases}}     \\ \hline 
		global $\beta_1$; global $\beta_3$      & 3.077                & 8.706          & 12.263              \\
		global $\beta_1$; individual $\beta_3$    & -3.15                & 6.741          & 13.938              \\
		individual $\beta_1$; individual $\beta_3$ & -3.66                & 5.456          & 14.516              \\
		individual $\beta_1$; global $\beta_3$    & 4.443                & 10.593         & 16.281             
	\end{tabular}
\end{table}

\subsubsection{Relative improvement over the modified mean model}

\begin{table}[H]
	\caption{Quantiles of improvement in prediction error of the proposed increment model over the zero model, obtained across all time intervals half as long as the entire time period.}
	\label{tab:supp_quantiles_resamples_zero}
	\begin{tabular}{lccc}
		\textbf{Increment model parameter} & \multicolumn{3}{c}{\textbf{Differences of squared error}} \\
		& \multicolumn{3}{c}{(zero model - increment model)}        \\
		& 1st Quartile        & Median        & 3rd Quartile        \\ \hline 
		\multicolumn{4}{l}{\textbf{global parameter} $\mathbf{\beta_2}$ \textbf{for prevalent cases}}     \\ \hline 
		global $\beta_1$; individual $\beta_3$                 & -18.295             & -1.368        & 4.112               \\
		individual $\beta_1$; global $\beta_3$                 & -0.467              & 3.545         & 6.599               \\
		global $\beta_1$; global $\beta_3$                     & -2.436              & 1.294         & 7.5                 \\
		individual $\beta_1$; individual $\beta_3$             & -4.197              & 2.283         & 8.006               \\ \hline 
		\multicolumn{4}{l}{\textbf{individual parameter} $\mathbf{\beta_2}$ \textbf{for prevalent cases}}     \\ \hline 
		global $\beta_1$; global $\beta_3$                     & 0.24                & 4.592         & 10.704              \\
		individual $\beta_1$; global $\beta_3$                 & 1.987               & 7.742         & 12.08               \\
		global $\beta_1$; individual $\beta_3$                 & -5.496              & 3.419         & 12.292              \\
		individual $\beta_1$; individual $\beta_3$             & -4.635              & 3.164         & 12.843             
	\end{tabular}
\end{table}

\subsubsection{Relative improvement over the linear regression model based on last-observation-carried-forward}

\begin{table}[H]
	\caption{Quantiles of improvement in prediction error of the proposed increment model over the linear regression model based on last-observation-carried-forward imputation, obtained across all time intervals half as long as the entire time period.}
	\label{tab:supp_quantiles_resamples_linreg}
	\begin{tabular}{lccc}
		\textbf{Increment model parameter} & \multicolumn{3}{c}{\textbf{Differences of squared error}} \\
		& \multicolumn{3}{c}{(linear regression - increment model}  \\
		& 1st Quartile        & Median         & 3rd Quartile       \\ \hline 
		\multicolumn{4}{l}{\textbf{global parameter} $\mathbf{\beta_2}$ \textbf{for prevalent cases}}     \\ \hline 
		global $\beta_1$; individual $\beta_3$                & 99.001              & 158.224        & 256.425            \\
		individual $\beta_1$; individual $\beta_3$            & 111.08              & 166.88         & 258.755            \\
		global $\beta_1$; global $\beta_3$                     & 108.934             & 167.659        & 273.855            \\
		individual $\beta_1$; global $\beta_3$                 & 111.081             & 168.264        & 276.345            \\ \hline 
		\multicolumn{4}{l}{\textbf{individual parameter} $\mathbf{\beta_2}$ \textbf{for prevalent cases}}     \\ \hline 
		global $\beta_1$; global $\beta_3$                     & 112.318             & 172.188        & 244.426            \\
		global $\beta_1$; individual $\beta_3$                 & 100.857             & 152.108        & 251.917            \\
		individual $\beta_1$; individual $\beta_3$             & 103.263             & 155.078        & 255.066            \\
		individual $\beta_1$; global $\beta_3$                 & 116.576             & 171.549        & 259.935           
	\end{tabular}
\end{table}

\newpage

\section{Validation: Recovering censored values}

\setlength\LTleft{0pt}
\setlength\LTright{0pt}
\begin{table}[H]
	\begin{tabular}{lcccc}
	\textbf{Prediction model} & \multicolumn{4}{c}{\textbf{Squared error}}       \\
	& Mean      & 1st Quartile & Median & 3rd Quartile \\ \hline 
	\multicolumn{5}{l}{\textbf{10\% of all reports censored at random}}          \\ \hline 
	Mean model                & 0.042     & 0.002        & 0.011  & 0.035        \\
	Modified mean model       & 0.042     & 0.002        & 0.011  & 0.035        \\
	Zero model                & 0.043     & 0.002        & 0.011  & 0.035        \\
	Linear regression         & 0.06      & 0.003        & 0.015  & 0.049        \\
	Increment model           & 0.043     & 0.002        & 0.009  & 0.03         \\ \hline 
	\multicolumn{5}{l}{\textbf{25\% of all reports censored at random}}          \\ \hline 
	Mean model                & 0.116     & 0.006        & 0.031  & 0.097        \\
	Modified mean model       & 0.115     & 0.006        & 0.03   & 0.094        \\
	Zero model                & 0.117     & 0.006        & 0.032  & 0.098        \\
	Linear regression         & 0.179     & 0.007        & 0.046  & 0.144        \\
	Increment model           & 0.13      & 0.004        & 0.024  & 0.085        \\ \hline 
	\multicolumn{5}{l}{\textbf{50\% of all reports censored at random}}          \\ \hline 
	Mean model                & 0.334     & 0.016        & 0.088  & 0.274        \\
	Modified mean model       & 0.333     & 0.016        & 0.087  & 0.273        \\
	Zero model                & 0.341     & 0.016        & 0.089  & 0.281        \\
	Linear regression         & 1.009     & 0.027        & 0.179  & 0.549        \\
	Increment model           & 0.315     & 0.011        & 0.062  & 0.219        \\ \hline 
	\multicolumn{5}{l}{\textbf{75\% of all reports censored at random}}          \\ \hline 
	Mean model                & 0.825     & 0.039        & 0.205  & 0.631        \\
	Modified mean model       & 0.823     & 0.039        & 0.205  & 0.63         \\
	Zero model                & 0.846     & 0.04         & 0.209  & 0.654        \\
	Linear regression         & 24582.387 & 0.156        & 1.499  & 7.024        \\
	Increment model           & 0.766     & 0.017        & 0.129  & 0.474       \\
	\end{tabular}
	\caption{Performance of the increment model with all parameters individually estimated and the benchmark models on recovering censored reports that were deleted at random at fixed rates of $10\%$, $25\%$, $50\%$ and $75\%$ in all $463$ DIVI hospitals with no missing data. For each hospital and each rate, the specific proportion of reports was randomly censored. All models were fit on the resulting data as described in Section 4.4 of the main text (i.e., censored observations were imputed with last-observation carried forward for the benchmark models and the same optimization was used for the increment model). We then calculated the mean squared squared error between the predicted and the true trajectory across the entire time interval. We repeated the censoring and fitting process $10$ times for each hospital and scenario and averaged the error. Here, we report the mean, median and 1st and 3rd quartile of these errors across all $463$ hospitals. \\}
	\label{tab:supp_leavesomeoutvalidation_mar}
\end{table}

\end{document}